\documentclass[12pt]{article}
\usepackage[letterpaper,margin=2cm]{geometry}
\usepackage{natbib}
\usepackage{algorithm}
\usepackage{algpseudocode}
\usepackage{adjustbox}
\usepackage{amsmath}
\usepackage{amssymb}
\usepackage{booktabs}
\usepackage{caption}
\usepackage[T1]{fontenc}
\usepackage{subcaption}
\usepackage{graphicx}
\usepackage{amsfonts}
\usepackage{hyperref}

\newcommand{\Tau}{\mathcal{T}}

\newcommand{\inp}{\mathbf{x}}

\newcommand{\weight}{\mathbf{W}}

\DeclareMathOperator*{\argmin}{arg\,min}
\DeclareMathOperator*{\expect}{\mathbb{E}}

\title{Subspace Adaptation Prior for Few-Shot Learning\thanks{Accepted at Machine Learning Journal, Special Issue of the ECML PKDD 2023 Journal Track}}

\author{Mike Huisman \and Aske Plaat \and Jan N. van Rijn}

\date{Leiden Institute of Advanced Computer Science, Leiden University}

\begin{document}

\maketitle

\begin{abstract}
Gradient-based meta-learning techniques aim to distill useful prior knowledge from a set of training tasks such that new tasks can be learned more efficiently with gradient descent.
While these methods have achieved successes in various scenarios, they commonly adapt \emph{all} parameters of trainable layers when learning new tasks.
This neglects potentially more efficient learning strategies for a given task distribution and may be susceptible to overfitting, especially in few-shot learning where tasks must be learned from a limited number of examples. 
To address these issues, we propose \emph{Subspace Adaptation Prior} (SAP), a novel gradient-based meta-learning algorithm that jointly learns good initialization parameters (prior knowledge) and layer-wise \emph{parameter subspaces} in the form of operation subsets that should be adaptable.
In this way, SAP can learn which operation subsets to adjust with gradient descent based on the underlying task distribution, simultaneously decreasing the risk of overfitting when learning new tasks.  
We demonstrate that this ability is helpful as SAP yields superior or competitive performance in few-shot image classification settings (gains between 0.1\% and 3.9\% in accuracy). 
Analysis of the learned subspaces demonstrates that low-dimensional operations often yield high activation strengths, indicating that they may be important for achieving good few-shot learning performance. 
For reproducibility purposes, we publish all our research code publicly.
\end{abstract}

\section{Introduction}

Humans are characterized by their ability to quickly learn new tasks and skills from only a limited amount of examples or experience. 
While deep neural networks are able to achieve great performance on various tasks \citep{krizhevsky2012imagenet,mnih2013playing, silver2016mastering, wurman2022outracing}, they require large amounts of data and compute resources to learn new tasks, restricting their success to domains where such resources are available.
One explanation for this gap in learning efficiency is that humans can efficiently draw on a large pool of prior knowledge and learning experience \citep{jankowski2011meta}, whereas deep neural networks are often trained from scratch or lack the appropriate prior.  

\emph{Meta-learning} \citep{schmidhuber1987evolutionary, thrun1998lifelong, naik1992meta, brazdil2022metalearning} is one potential solution to this problem as it can distill a good prior from a set of past learning experiences that facilitates efficiently learning new tasks. 
Model-agnostic meta-learning (MAML) \citep{finn2017model} is a popular gradient-based meta-learning algorithm that learns a prior in the form of the initialization parameters of the network.
Learning a new task is then done by performing gradient descent starting from this meta-learned initialization. 
This approach, which is also widely used by techniques that are based on MAML \citep{lee2018gradient, Flennerhag2020Meta-Learning, parkO19metacurvature, yoon2018bayesian, nichol2018reptile}, updates all of the parameters of every \emph{trainable} layer with gradient descent when learning new tasks, which may be suboptimal for a given task distribution and may lead to overfitting since there are more degrees of freedom to fit the noise in the data.
Especially in few-shot learning, where tasks are noisy due to the fact that only limited examples are available, these issues could hinder performance.  

To address these issues and investigate the research question of whether the few-shot learning performance of deep neural networks can be improved by meta-learning which subsets of parameters to adjust, we propose a new gradient-based meta-learning technique called \emph{Subspace Adaptation Prior} (SAP) that jointly learns good initialization parameters as well as layer-wise subspaces in which to perform gradient descent when learning new tasks. 
More specifically, SAP is given access to a candidate pool of \emph{operations} for every layer that transforms the hidden representations, and it learns which of these subsets to adjust in order to learn new tasks quickly, similar to DARTS~\citep{liu2018darts}. 
Here, every operation corresponds to a parameter subspace. 
Note that this method serves as a form of regularization and allows SAP to find more efficient adaptation strategies than adjusting all parameters of trainable layers.
In addition, it utilizes implicit gradient modulation to warp \citep{lee2018gradient, Flennerhag2020Meta-Learning} these subspaces per layer such that gradient descent can quickly adapt to new tasks, if they share a common structure.

We empirically demonstrate that SAP is able to find efficient parameter subspaces, or operation subsets, that match the underlying task structure in simple synthetic settings and yield good few-shot learning. 
Moreover, SAP outperforms gradient-based meta-learning techniques---that do not have the ability to learn in which structured subspaces to perform gradient descent---on few-shot sine wave regression and performs on-par or favorably in various few-shot image classification settings. 
In short, our contributions are the following:
\begin{itemize}
    \item We propose SAP, a new meta-learning algorithm for few-shot learning that jointly learns good initialization parameters and parameter subspaces in the form of operation subsets in which to perform gradient descent.
    \item We demonstrate the advantage of learning parameter subspaces as SAP outperforms existing methods by at least 18\% on few-shot sine wave regression and yields competitive or superior performance on popular few-shot image classification benchmarks (improvements in classification accuracy scores range from 0.1\% to 3.9\%).
    \item We investigate the learned layer-wise parameter subspaces on synthetic few-shot sine wave regression and image classification problems and find that small subsets of adjustable parameters (simple parameter subspaces), including feature transformations such as element-wise scaling and shifting are assigned large weights, suggesting that they play an important role in achieving good performance with SAP.
    \item For reproducibility and verifyability purposes, we make all our research code publicly available.\footnote{See: \url{https://github.com/mikehuisman/subspace-adaptation-prior}}
\end{itemize}

\section{Related work}

We review related work on optimization-based meta-learning, neural architecture search and gradient modulation.

\paragraph{Optimization-based meta-learning}
Our proposed technique belongs to the category of optimization-based meta-learning \citep{vinyals2017talk, huisman2021}, which employs optimization methods to learn new tasks \citep{yoon2018bayesian, Bertinetto19, lee2019meta}.
These methods aim to meta-learn good settings for various hyperparameters, such as the initialization parameters, such that new tasks can be learned quickly using optimization methods. 
These methods vary from regular stochastic gradient descent, as used in MAML \citep{finn2017model} and Reptile \citep{nichol2018reptile}, to meta-learned procedures where a network updates the weights of a base-learner \citep{ravi2017optimization, andrychowicz2016learning, li2017metasgd, rusu2018meta, li2018learning, huisman2022stateless}. 
SAP aims to learn good initialization parameters such that new tasks can be learned quickly with regular gradient descent, similar to MAML. 

This is a form of transfer learning \citep{taylor2009transfer, pan2009survey} where we transfer knowledge---in this case the initialization parameters---obtained on a set of source tasks to a new target task that we are confronted with.
The idea is also related to the idea of domain adaptation (DA) \citep{daume2009frustratingly, farahani2021brief}, although in DA it is often assumed that we have a single task but two different data distributions (a source distribution and a target distribution). 
Note that in deep meta-learning \citep{huisman2021, hospedales2020meta}, we have set of various different training tasks and aim to transfer knowledge to a new target task, different from the ones seen at training time.

\paragraph{Neural architecture search (NAS) for meta-learning}

The techniques mentioned above assume a pre-specified network architecture.
Recently, there has been some work on combining meta-learning with neural architecture search, where the architecture can also be learned.
\citet{kim2018auto} performs meta-learning as a subroutine to NAS, meaning that meta-training is performed for every candidate architecture, which can be computationally expensive. 
This problem can be overcome by combining gradient-based meta-learning with gradient-based neural architecture search such that the architecture and initialization parameters can be optimized jointly instead of separately. 
A popular gradient-based meta-learning algorithm is DARTS \citep{liu2018darts} which starts with a candidate pool of operations (as in SAP) and learns which of them to use, thereby learning an appropriate architecture. 
Learning which subspaces or subsets of operations to use per layer, as done in SAP, can be seen as applying DARTS over the candidate operation sets. 
A difference between DARTS and SAP is that we fix the base-learner parameters when adapting to new tasks, which can then serve to warp, or transform, the gradients such that gradient descent can quickly move to a good solution for new tasks (see below).
Moreover, SAP updates the initialization parameters of all meta-trainable parameters with a MAML-like update (to maximize post-adaptation performance), while DARTS uses a Reptile-like update (to maximize multi-step performance).
We describe DARTS in full detail in Section~\ref{sec:darts}.

\citet{lian2019towards} were the first to combine DARTS \citep{liu2018darts} with gradient-based meta-learning in order to learn a base-learner architecture that can be quickly adapted to new tasks. 
They perform hard-pruning, which requires re-running the meta-training phase for every new task, which is computationally expensive. 
In parallel to this work, \citet{elsken2020meta} proposed a similar approach (MetaNAS) that does not perform hard-pruning and thus side-steps these expensive re-running procedures.
In contrast to these works, which learn and adapt the base-learner network architecture as well as all of the parameters to every new task, SAP assumes a fixed base-learner architecture as a starting point and aims to learn a set of operations that are inserted per layer (see Section~\ref{sec:theory}) that are responsible for quickly adapting to new tasks.
In SAP, the architecture of the network is frozen at test time, in contrast to, for example, the architecture of the networks learned by MetaNAS \citep{elsken2020meta}.

\paragraph{Gradient modulation in gradient-based meta-learning}
Recent works that build upon MAML have shown that gradient modulation can improve the generalization of optimization-based techniques \citep{sun2019meta}.
Explicit gradient modulation techniques directly transform the gradient updates when learning new tasks \citep{simon2020modulating} through, for example, diagonal matrix multiplication \citep{li2017metasgd}, or block-diagonal preconditioning \citep{parkO19metacurvature}.  
Implicit gradient modulation techniques do not directly operate on the gradients but rely on indirect transformations. 
CAVIA \citep{zintgraf2019fast} separates shared parameters from context parameters.
The latter serve as additional inputs to one or more layers of the neural network and are adjusted when learning a new task, whilst the shared parameters are kept fixed. 
Other examples of implicit gradient modulation methods are T-Net \citep{lee2018gradient} and Warp-MAML \citep{Flennerhag2020Meta-Learning}.
SAP also performs implicit gradient modulation in a similar fashion to these two techniques.  

T-Net inserts linear projection transformations directly after every matrix multiplication in the base-learner. 
The weights of these transformations are frozen when learning new tasks, and only the base-learner weights are adjusted. 
The goal is to meta-learn good initialization parameters of the base-learner weights as well as the transformation weights, such that new tasks can be learned more quickly. 
These transformation layers serve to implicitly modulate the gradients of the base-learner parameters so that gradient descent can quickly move to good solutions for new tasks. 
MT-Net is an extension to T-Net, which also learns to mask certain features, preventing them from being adapted when learning new tasks. 
We also investigated whether feature masking was useful for SAP, but found that it decreased performance. 
Warp-MAML is a generalization of T-Net as it does not require that the inserted transformation layers are linear, that is, the theoretical framework allows these transformation layers to be non-linear and consist of multiple layers (arbitrary neural networks).

Both T-Net and Warp-MAML adjust all parameters of trainable layers, as is common in gradient-based meta-learning.
However, this may be suboptimal for a given task distribution and lead to overfitting due to the large degree of freedom to fit the noise in the data. 
MT-Net, on the other hand, freezes certain features, which, in turn, also requires certain weights to be frozen but this is rather inflexible as that does not allow us to perform simple operations such as element-wise scaling of all features, which may be helpful for a given task distribution. 
To overcome these issues, we propose SAP, which learns per trainable layer which operations from a pre-defined candidate pool to use and adapt when learning new tasks, instead of resorting to regular matrix multiplications in which all weights are adjusted when learning new tasks (as done by other methods).
While the expressivity of SAP is equivalent to T-Net and Warp-MAML (when using linear warp layers), the candidate pool of operations allows SAP to learn which operations are important for the given task distribution, thereby structuring the weight updates. 

SAP is similar to T-Net and Warp-MAML in the sense that the linear base layers $\weight^\ell$ (see Section~\ref{sec:theory}) of the network in SAP can be seen as the warp layers or transformation layers that are used in T-Net and Warp-MAML, which act as implicit preconditioning layers that warp the loss surface to aid gradient descent in finding a good solution. 
Due to the similarities between T-Net, Warp-MAML, and SAP, they serve as excellent baselines to investigate whether the ability of SAP to learn which operation subsets to adapt when learning new tasks is helpful for few-shot learning. 
Concurrently to our work, \citet{jiang2022subspace} have proposed a subspace meta-learning algorithm. 
Whilst the title is similar, they explicitly meta-learn the bases for $K$ subspaces.
Then, when learning a new task, they aim to find linear combinations of the basis vectors of each of the subspaces that give rise to the best parameters for the given task in the subspaces.
The subset containing the parameters with the lowest training loss is then used to obtain predictions for the query/test set. 
Note that their work is different in that we do not learn basis vectors for different subspaces, but instead insert candidate operations that act to transform intermediate representations in the base-learner network to allow for faster learning and modulating the gradients.

\section{Preliminaries}
In this section, we introduce the problem setup and notation that we will use throughout the paper. 

\subsection{Few-shot meta-learning}
\label{sec:backgroundfsl}

In few-shot learning \citep{lu2020learning, wang2020generalizing, bendre2020learning}, the goal is to learn a new task $\Tau_j$ from a limited number of examples. 
Every task $\Tau_j = \{ D^{tr}_{\Tau_j}, D^{te}_{\Tau_j} \}$ consists of a \emph{support set} $D^{tr}_{\Tau_j}$ that is used for learning the new task and a \emph{query set} $D^{te}_{\Tau_j}$ for evaluating how well the task was learned. 
Learning new tasks with deep neural networks from limited amounts of data is challenging.
Meta-learning aims to overcome this challenge by learning how to learn on a distribution of training tasks $p_{\mathit{train}}(\Tau)$ in the hope that new tasks (not seen during training) from a similar distribution can be learned more efficiently.

Meta-learning is often done in three stages. 
In the \textit{meta-training} stage, the meta-learner is presented with training tasks and uses them to adjust the prior, such as the initialization parameters.
After every pre-determined number of training tasks, the \textit{meta-validation} stage takes place, where the learner is validated on unseen meta-validation tasks.
Finally, after the training is completed, the learner with the best validation performance is evaluated in the \textit{meta-test} phase, where the learner is confronted with new tasks that have not been seen during training and validation. 
Importantly, the tasks between the meta-training, meta-validation, and meta-test phases are disjoint. 
For example, in image classification, the classes in the meta-training tasks are not allowed to occur in meta-test tasks as we are interested in measuring the learning ability instead of memorization ability. 

In \emph{$N$-way $k$-shot classification} \citep{finn2017model, vinyals2016matching, snell2017prototypical}, the support set $D^{tr}_{\Tau_j}$ of every task $\Tau_j$ contains $N$ classes and exactly $k$ shots, or equivalently, examples, per class, thus $|D^{tr}_{\Tau_j}| = k \cdot N$. 
Moreover, the query set $D^{te}_{\Tau_j}$ contains unseen examples from the same $N$ classes, so that it can be evaluated how well the concepts in the support set have been learned. 
For regression problems, there is no notion of classes, but the same setup can be used, i.e., support sets consist of $k$ shots of one regression function and the query sets of unseen examples of that same regression function.

\subsection{Model-agnostic meta-learning (MAML)}
A popular gradient-based meta-learning technique is model-agnostic meta-learning, or MAML \citep{finn2017model}, which we briefly review here. 
MAML aims to learn good initialization parameters $\mathbf{\theta}$ of a neural network $f_{\mathbf{\theta}}$ such that new tasks can be learned in a few gradient update steps from that initialization.

This initialization is obtained by interleaving inner- and outer-update steps during the meta-training phase. 
At the \emph{inner-level}, the model $f_{\mathbf{\theta}}$ is presented with a task $\Tau_j$, which it aims to learn by making $T$ gradient update steps on the support set of that task $D^{tr}_{\Tau_j}$, that is,  
\begin{align}
    \mathbf{\theta}^{(t+1)}_j = \mathbf{\theta}^{(t)} - \alpha \nabla_{\mathbf{\theta}^{(t)}} \mathcal{L}_{D^{tr}_{\Tau_j}}(\mathbf{\theta}^{(t)}),\label{eq:graddesc}
\end{align} where $\alpha$ is the \emph{inner} learning rate and $\mathcal{L}_{D^{tr}_{\Tau_j}}(\mathbf{\theta}^{(t)})$ the loss of the network with parameters $\mathbf{\theta}^{(t)}$ on the support set of task $\Tau_j$ at time step $t$. 
Before learning a task, $\theta^{(0)}$ is initialized as $\theta$. 
These task-specific parameters $\mathbf{\theta}^{(t)}_j$ are then used to evaluate how well the task was learned. 
This loss signal is then propagated backward to the initialization parameters $\mathbf{\theta}$ to compute the update direction.
The latter corresponds to \emph{outer-level} learning: adjusting the initialization parameters over a single task, or more generally, a batch of tasks $B$ on which the inner-level update steps were made
\begin{align}
    \mathbf{\theta} = \mathbf{\theta} - \beta \nabla_{\mathbf{\theta}} \sum_{\Tau_j \in B} \mathcal{L}_{D^{te}_{\Tau_j}}(\mathbf{\theta}^{(T)}_j),
\end{align}
where $\beta$ is \emph{outer} learning rate.
This update requires the computation of second-order gradients as we have to compute a gradient of a gradient, which is expensive as it has a complexity quadratic in the number of parameters.
This can be sidestepped by using a first-order approximation. 
Importantly, note that the inner-level updates are based on the loss on the support set while the outer-level updates are based on the loss on the query set after adaptation, stimulating generalization. 
For simplicity, the gradient update rules are shown in the case that a single update is made per task, even though the idea generalizes to the case of multiple updates per task. 

MAML has been proven to be effective at learning new tasks from limited amounts of data \citep{finn2017model} as well as capable of approximating any learning algorithm \citep{finn2018meta} by means of selecting a proper initialization $\mathbf{\theta}$, under the assumption that the used network is ``sufficiently'' deep.

\begin{figure}[htb]
    \centering
    \includegraphics[scale=0.7]{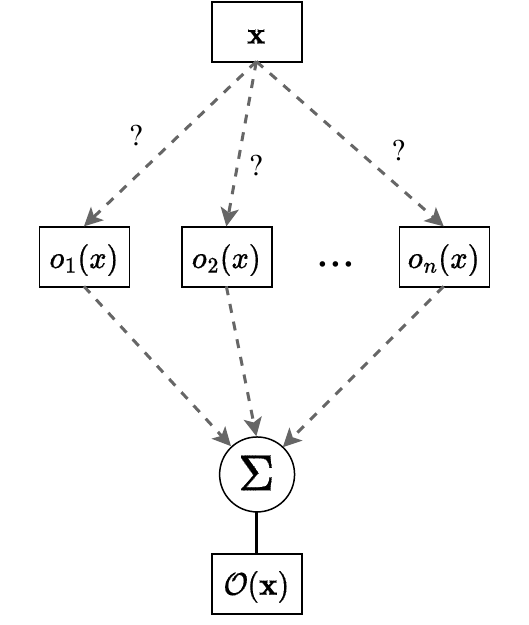}
    \caption{Intuitive visualization of DARTS. It is given a set of candidate operations $\mathcal{O}$ and aims to learn the weights of the edges (indicated as ?), corresponding to the strengths of the different operations $o_i(\mathbf{x})$. The output of the weighted graph is a convex combination of the different operations $\mathcal{O}(\mathbf{x}) = \sum_{i=1}^{n} w_i o_i(\mathbf{x})$.}
    \label{fig:darts}
\end{figure}

\subsection{Differentiable neural architecture search (DARTS)}
\label{sec:darts}

DARTS \citep{liu2018darts} is a gradient-based neural architecture search method, where the goal is to find a suitable neural architecture for a given problem.
To do this, DARTS assumes a set of candidate operations that can be used to transform an input into an output. 
These candidate operations form a weighted graph as shown in Figure~\ref{fig:darts}.
In the figure, every node $o_i(\mathbf{x})$ corresponds to a candidate operation and the weights of the edges correspond to the activation strengths of the different operations.
These weights are initially unknown and DARTS aims to learn them jointly with the initial parameters of every operation. 
The output of the layer in the figure is given by 
\begin{align}
    \mathcal{O}(\mathbf{x}) = \sum_{i=1}^{n} w_i o_i(\mathbf{x}),
\end{align}
where $w_i$ is the weight of operation $i$ and $\sum_{i=1}^{n} w_i = 1$ (e.g., by using a softmax).
For our purposes, we only consider DARTS for searching over operations for a single layer, but it can be used for multi-layer architectures as well.

In addition to learning the weights $w_i$, DARTS simultaneously learns good parameters for every operation $o_i$.
We denote the group of all activation weights as $\lambda = \{ w_1, w_2,\ldots,w_n \}$ and all operation parameters as $\theta$.
DARTS adopts a method similar to MAML for learning $\lambda$ and $\theta$. 
That is, given a training task $\Tau_j = (D^{tr}_{\Tau_j}, D^{te}_{\Tau_j})$, DARTS performs a gradient update step on the operation weights $\theta$ as follows
\begin{align}
    \mathbf{\theta}'_j = \mathbf{\theta} - \alpha \nabla_{\mathbf{\theta}, \lambda} \mathcal{L}_{D^{tr}_{\Tau_j}}(\mathbf{\theta}, \lambda).
\end{align}
Note that this is similar to Equation~\ref{eq:graddesc} with the exception that we now have activation strength parameters $\lambda$, which are kept constant during this inner-loop adaptation step.
After updating the operation parameters $\theta$, DARTS computes the loss of the new model on the query set, i.e., $\mathcal{L}_{D^{te}_{\Tau_j}}(\theta'_j,\lambda)$ and updates the activation strengths using gradient descent on this loss
\begin{align}
    \lambda  = \lambda - \beta \nabla_{\lambda} \mathcal{L}_{D^{te}_{\Tau_j}}(\theta'_j,\lambda).
\end{align}
Similarly to MAML, this update also contains second-order gradients, but first-order approximations can be made. 
In DARTS, the weights of the operations $\theta$ are simply updated to their new values, that is, $\theta = \theta_j'$, i.e., after every task in the meta-train set, we update the initialization parameters $\theta$ to the parameters that were obtained after training on task $\Tau_j$.

\begin{figure}[htb]
    \centering
    \begin{subfigure}{0.38\textwidth}
    \includegraphics[width=\linewidth]{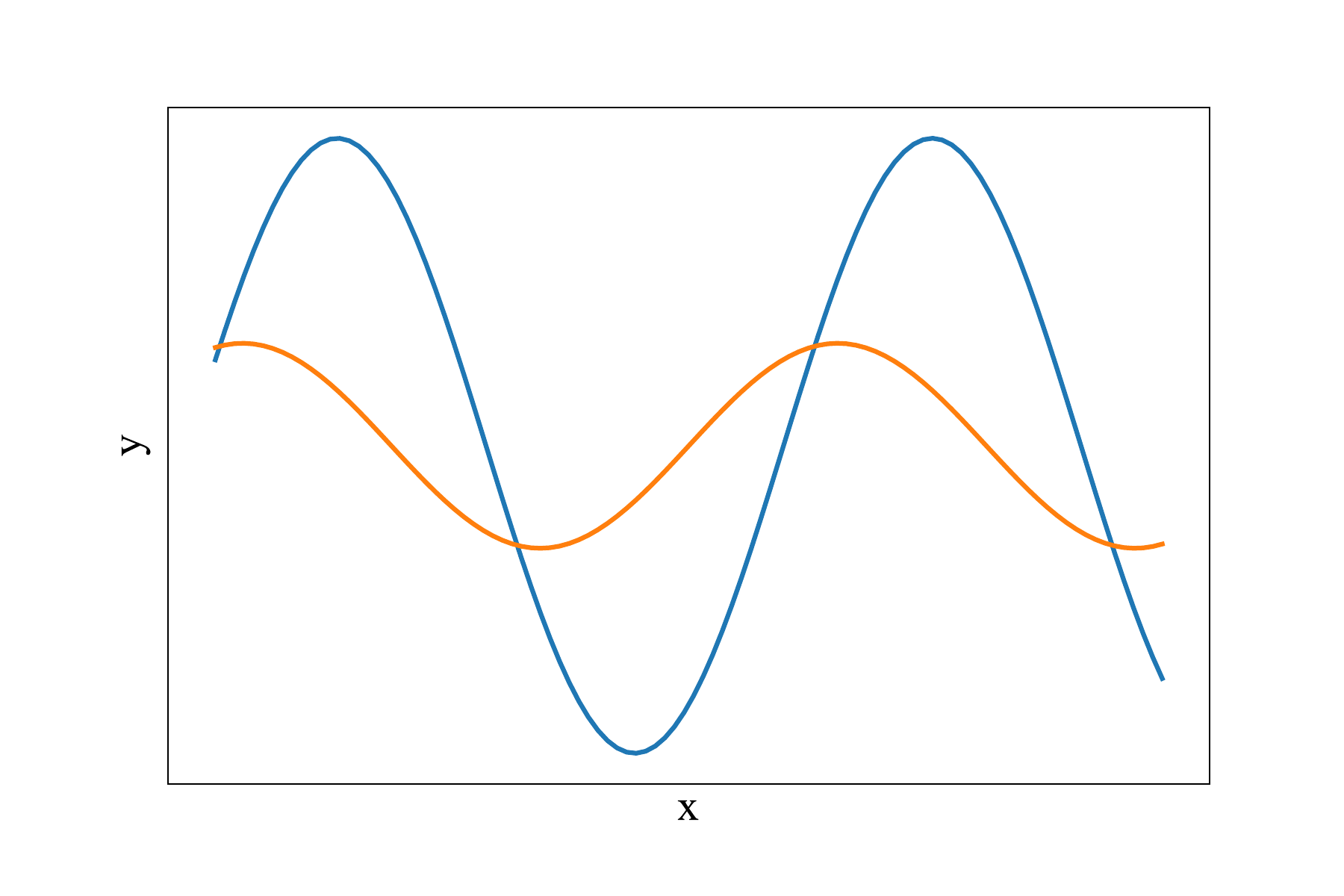}
    \caption{Sine wave tasks}
    \end{subfigure}
    \begin{subfigure}{0.6\textwidth}
    \includegraphics[width=\linewidth]{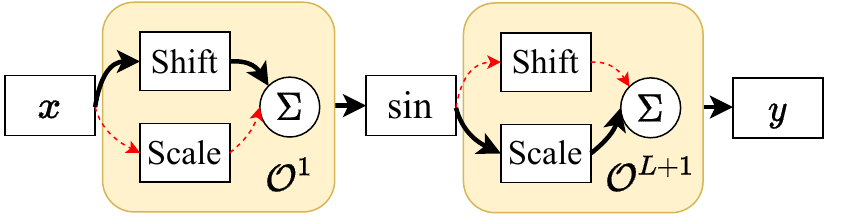}
    \caption{Learned subspaces}
    \end{subfigure}
    \caption{SAP can learn the activation strengths of candidate operations $\mathcal{O}^\ell$ (corresponding to parameter subspaces) that match the problem structure. Suppose we are given a sine wave task distribution, where every task $\Tau_j$ is a sine wave $g_j(x) = A_j\cdot \sin(x-p_j)$, where $p_j$ is the phase and $A_j$ the amplitude. Instead of adapting all parameters of the network on a new task, SAP can learn to keep the sine network parameters (sin) frozen and that the input shift (shift in $\mathcal{O}^1$) and output scale (scale in $\mathcal{O}^{L+1}$) are the most important operations to adjust (bold and dark-colored arrows), matching the role of the phase and amplitude, respectively. }
    \label{fig:sapidea}
\end{figure}

\section{Subspace Adaptation Prior}
\label{sec:theory}

In this section, we motivate and present our proposed technique called \emph{Subspace Adaptation Prior} (SAP).

\subsection{Intuition and operations}\label{sec:operations}

Our method (SAP) builds on MAML as we also aim to learn good initialization parameters such that good performance can be achieved after a small number of gradient updates.
However, MAML adapts all of its network parameters when presented with a new task, which may be suboptimal for the given task distribution and lead to overfitting.
Our method, SAP, is given a pool of candidate operations per layer (described below) and it learns per layer which subset of operations should be adjusted to adapt to a new task. 
Since all of the operations that SAP can choose from per layer are subsumed in terms of expressivity by a full-rank matrix multiplication (or convolution in the case of image data), this can be understood as learning in which {\it parameter subspaces} to perform gradient descent so that new tasks can be learned more efficiently.

This is a form of regularization and can help the network to exploit structures in problems.
For example, take the distribution of tasks $\Tau_j$ corresponding to different sine waves $g_j(x) = A_j\cdot \sin(x-p_j)$, where $A_j$ is the amplitude and $p_j$ the phase.
There exists a common structure amongst these tasks: a given sine wave can be transformed into any other sine wave by simply shifting the input and scaling the output. 
This has been visualized in Figure~\ref{fig:sapidea}.
Techniques that adapt all parameters may overwrite the sine function and overfit to the noise, whereas theoretically, SAP could learn to keep these parameters fixed and that shifting the input and scaling the output are the most important operations and consequently, that gradient descent should be performed in the parameter subspaces corresponding to these operations.
Sine waves form a simplistic example to demonstrate the idea of SAP, however, we note that also for image classification tasks, simple operations such as scaling and shifting feature maps can be useful too \citep{sun2019meta, perez2018film, requeima2019fast}.
SAP can discover such underlying structures and use them to enhance its few-shot learning abilities.

\paragraph{Candidate operations}

The candidate operations that SAP uses are specified by hand before meta-training.
In order to preserve the original expressivity of the base-learner network, the operations are elementary linear algebra operations that are subsumed by full-rank matrix multiplication.  

Table~\ref{tab:pool} displays all the operations that we use for both fully-connected and convolutional layers. 
The MTL scale operation was proposed by \citet{sun2019meta}.
By construction, we require that the output of an operation set must have the same dimensionality as the input. 
Recall that in the case of fully-connected layers, all candidate operations can be expressed by a single matrix multiplication where only a subset of the entries is used.
For example, an element-wise scale can be performed by multiplying the input with a diagonal matrix where the diagonal entries correspond to the element-wise scalars, and the non-diagonal entries are zero. 
In this way, every candidate operation occupies a \emph{part} of the full operation set matrix.  
This also holds for convolutions, which can be seen as a stack of matrices.

\begin{table}[htb]
\centering
\begin{adjustbox}{width=\linewidth}
\begin{tabular}{@{\extracolsep{8pt}}llll}
\toprule 
\multicolumn{2}{c}{\textbf{Fully-connected}}   & \multicolumn{2}{c}{\textbf{Convolutional}}               \\
\cline{1-2} \cline {3-4}
Operation & Dimensionality & Operation & Dimensionality \\
\midrule
Identity & N.A. & Identity & N.A. \\
Matrix multiplication & $d \times d$  & Convolution & $C \times C \times k \times k$ \\
SVD-matrix multiplication & $d \times v$ & SVD convolution & $C \times C \times k \times v$       \\
Element-wise scale & $d$ & 1x1 convolution & $C \times C$    \\
Scalar scale & $1$  &  MTL scale & $C \times C$  \\
Vector shift & $d$  & Channel-wise scale & $C$ \\
Scalar shift& $1$ & Channel-wise shift & $C$ \\
                   & & Scalar shift  & $1$      \\
\bottomrule
\end{tabular}
\end{adjustbox}
\caption{The candidate operations for fully-connected and convolutional network layers and the corresponding dimensionality of the subspace in which gradient will be performed. Here, $d$ is the dimensionality of the input in the case of a fully-connected layer and $C$ is the number of input and output dimensions of candidate operations in the case of convolutional layers. $k$ is the kernel size of convolutions and $v < k$ is a variable dimension for SVD matrices.
\label{tab:pool}}
\end{table}

We also include singular value (SVD) decomposition operations, where three $v$-rank matrices $A = U\Sigma V^T$ are multiplied to obtain a transformation matrix $A \in m \times n$ with the same dimensionality as a full-rank transform $T \in \mathbb{R}^{m \times n}$ (although with a lower rank). 
Here, $U \in \mathbb{R}^{m \times v}$, $\Sigma \in \mathbb{R}^{v \times v}$, and $V^T \in \mathbb{R}^{v \times n}$.
The obtained transformation $A$ is then applied to the input.

Below, we describe how these operations are interleaved with the base-learner network and how SAP learns which subsets to adjust. 

\subsection{The algorithm}
\label{sec:sapalgorithm}

\begin{figure}[!hb]
    \centering
    \includegraphics[scale=0.5]{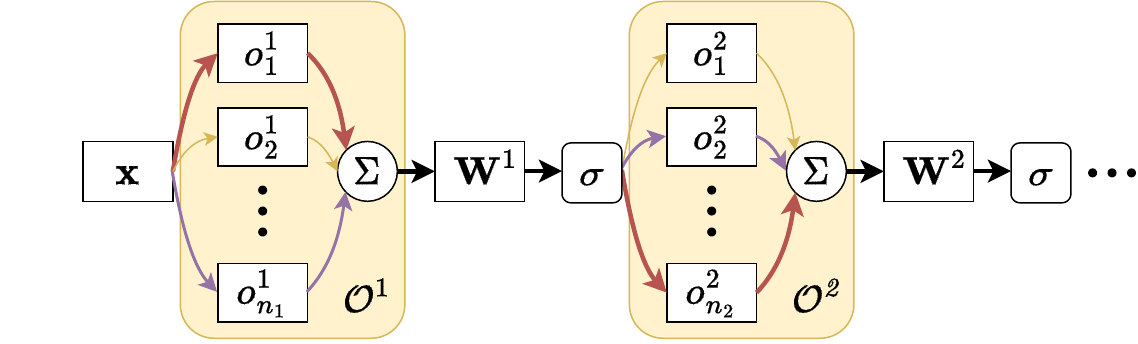}
    \caption{A diagram of a feed-forward pass in SAP. Sets of operations $\mathcal{O}^\ell$ are interleaved with base-learner weights $\weight^\ell$. 
    The operation sets perform a convex combination of a number of operations $\{ o^\ell_1,\ldots,o^\ell_{n_\ell} \}$. SAP learns the strengths of each of the candidate operations and thereby learns in which parameter subspaces gradient descent can effectively adapt the network to learn new tasks. The operation strengths and the weight matrices $\weight^\ell$ are frozen when adapting to new tasks. Only the operation parameters are adjusted at test time.}
    \label{fig:sap}
\end{figure}

\paragraph{Architecture}
Let $f_{\mathbf{\theta}}$ be a neural network with parameters $\mathbf{\theta}$, where the output, or prediction, is given by
\begin{align}
    f_\mathbf{\theta}(\inp) =  \weight^L \sigma( \ldots \sigma(\weight^2 \sigma(\weight^1 \inp))).
\end{align}
Here, $L$ is the number of layers of the network, $\sigma$ is a non-linear activation function, and $\weight^{\ell}$ is the weight matrix for layer $\ell \in \{ 1,2,\ldots,L \}$ (which can also include the bias by concatenating a 1 at the top of the input vector).
Note that $\theta = \{ \weight^1, \weight^2,\ldots, \weight^L \}$ is the set of all base-learner weight matrix parameters. 
In SAP, we insert sets of candidate operations $\mathcal{O}^\ell = \{ o^\ell_1,\ldots,o^\ell_{n_\ell}  \}$ before the application of weight matrices $\weight^\ell$ and after computing the final output, as shown in Figure~\ref{fig:sap}.
Here, $n_\ell$ is the number of operations in the candidate set $\mathcal{O}^{\ell}$ in layer $\ell$. 
Each of these operations $o^\ell_i \in \mathcal{O}^{\ell}$ act on the input, giving rise to partial outputs $o^\ell_i(\mathbf{z}^\ell)$ of the same dimensionality of the inputs, where $\mathbf{z}^\ell$ is the input to the $\ell$-th operation layer.
The final output of applying the candidate operations is a convex combination of the partial outputs, that is,
\begin{align}
    \mathcal{O}^\ell \mathbf{z}^\ell = \sum_{i=1}^{n_\ell} w^\ell_i o^\ell_i(\mathbf{z}^\ell),
\end{align}
where $\mathbf{z}^1 = \inp$ and $w^\ell_i$ is the activation strength of operation $o^\ell_i$. We require that $\sum_{i=1}^{n_\ell} w^\ell_i = 1$ and $0 \leq w_i \leq 1$.  
Learning these activation strengths can be seen as neural architecture search.  
Thus, the output of the neural network in SAP is given by 
\begin{align}
    f_\mathbf{\Theta}(\inp) = \mathcal{O}^{L+1} \weight^L \mathcal{O}^L  \sigma( \ldots \sigma( \weight^2 \mathcal{O}^2  \sigma(\weight^1 \mathcal{O}^1 \inp))), \label{eq:forward}
\end{align} where $\mathbf{\Theta} = \{ \mathbf{\theta}, \mathbf{\phi}, \mathbf{\lambda}  \}$ is the set of the initial hyperparameter values for the base-learner weights ($\theta$), the operation weights ($\phi$), and the activation weights ($\lambda$). 
Note that $\phi = \{ \mathcal{O}^1,\mathcal{O}^2,\ldots,\mathcal{O}^{L+1} \}$ are the parameters corresponding to the operations in all layers (see Section~\ref{sec:operations}), and $\lambda$ is the set containing all $w_i^\ell$ for all layers $\ell \in \{ 1,2,\ldots,L+1 \}$.

Importantly, each of these candidate operations $o^\ell_i$ are \emph{subsumed} or equivalent in terms of expressivity with full-rank matrix multiplication. 
For example, candidate operations can include element-wise shifting or multiplication of the input by a fixed scalar or by a vector, which can also be done by weight matrix multiplication. 
Since the application of a set of operations $\mathcal{O}^{\ell}$ of such expressivity can be seen as a single matrix multiplication (hence the suggestive notation), {\it the expressivity of an SAP network is equivalent to that of the original network.}
To see this, note that the application of two weight matrices to an input can be written as the application of a single weight matrix to the input $\inp$, that is, $\weight (\mathcal{O} \inp) = (\weight \mathcal{O}) \inp = \weight' \inp$, where $\weight$ and $\mathcal{O}$ are weight matrices, and $\weight' = \weight \mathcal{O}$.
For the sake of another example, suppose that we have a set of two operations in $\mathcal{O}$: scalar multiplication $s \cdot \mathbf{z}$ and matrix multiplication $\mathbf{M} \mathbf{z}$ (preserving the dimensionality of $\mathbf{z}$).
Furthermore, suppose that the two operations are applied with activation strengths $w_1$ and $w_2$, granting us the output $\mathbf{z}' = w_1 s \cdot \mathbf{z} + w_2 \mathbf{M} \mathbf{z}$.
We can rewrite this as $\mathbf{z}' = w_1 sI \mathbf{z} + w_2 \mathbf{M} \mathbf{z} = (w_1 s I + w_2 \mathbf{M}) \mathbf{z} = \mathcal{O} \mathbf{z}$, where $\mathcal{O} = (w_1 s I + w_2 \mathbf{M})$ and $I$ is the identity matrix.
For a more intuitive example, suppose that a base-layer is a fully-connected layer, and we add a fully-connected operation to alter the resulting representation, maintaining the original dimensionality.
The composition of the two fully-connected layers is effectively linear and equally expressive as a single fully-connected layer. 
\textbf{Thus, introducing the operations used by SAP does not alter the expressivity of the original base-learner network.}

Crucially, this insight that we can write the weighted combination of different operations as a single weight matrix multiplication $\mathcal{O} \inp$, where $\mathcal{O}$ is a weighted combination of different structured matrices, reveals that SAP effectively learns what subset of parameters of this weight matrix $\mathcal{O}$ and thus of $\weight \mathcal{O}$ to adjust by learning the activation strengths $\lambda$. In this work, we use the expressions ``learning which subsets of parameters to adjust'' and ``learning in what subspaces to perform gradient descent'' synonymously.

\paragraph{Meta-learning}

The activation strengths $w_i^\ell$ are meta-learned by SAP in addition to the initialization parameters of the operations $\mathcal{O}^\ell$ and the base-learner weights $\weight^\ell$.
Note that learning the $w_i^\ell$ corresponds to learning in which parameter subspaces gradient descent is performed when learning new tasks, which can be done through the layer-wise application of the gradient-based neural architecture search technique DARTS \citep{liu2018darts}.  
Let $\mathbf{\theta}$ denote the initial parameters of the weight matrices $\weight^\ell$, $\mathbf{\phi}$ the parameters of all candidate operations $\mathcal{O}^\ell$, and $\mathbf{\lambda}$ the activation strengths $w^\ell_i$ of all individual candidate operation.
Recall that $\mathbf{\Theta} = \{ \mathbf{\theta}, \mathbf{\phi}, \mathbf{\lambda}  \}$.

When presented with a new task $\Tau_j$, the candidate operation activation strengths $\mathbf{\lambda}$ and the base-learner parameters $\mathbf{\theta}$ are frozen, and only the candidate operation parameters $\mathbf{\phi}$ are updated using gradient descent for $T$ steps
\begin{align}
    \mathbf{\phi}^{(t+1)}_j \leftarrow \mathbf{\phi}^{(t)} - \alpha \nabla_{\mathbf{\phi}^{(t)}} \mathcal{L}_{\Tau_j}(\mathbf{\theta}, \mathbf{\phi}^{(t)}, \mathbf{\lambda}), \label{eq:graddescent}
\end{align} where $\mathbf{\phi}^{(0)}$ is initialized with $\mathbf{\phi}$.
At the meta-level, the goal is to find good initial parameter settings for all involved parameters such that the task-specific performance is maximized. 
Thus, we wish to find
\begin{align}
    \argmin_{\mathbf{\Theta} = \{ \mathbf{\theta}, \mathbf{\phi}, \mathbf{\lambda}  \}} \expect_{\Tau_j \backsim p(\Tau)} \mathcal{L}_{\Tau_j}(\mathbf{\theta}, \mathbf{\phi}^{(T)}_j, \mathbf{\lambda}),\label{eq:meta-objective}
\end{align} where $\mathbf{\phi}^{(T)}_j$ denotes the task-specific parameters obtained through one or more gradient update steps on task $\Tau_j$. 
In other words, we wish to find good initial values for the parameters $\theta$, $\phi$, and $\lambda$ such that new tasks can be learned quickly by updating the operation parameters $\phi$.
This meta-objective can also be optimized through gradient descent by updating
\begin{align}
    \mathbf{\Theta} \leftarrow \mathbf{\Theta} - \beta \nabla_{\mathbf{\Theta}} \sum_{\Tau_j \in \mathcal{B}} \mathcal{L}_{\Tau_j}(\mathbf{\theta}, \mathbf{\phi}^{(T)}_j, \mathbf{\lambda}). \label{eq:outerobjective}
\end{align}

The full algorithm for application to few-shot learning is shown in Algorithm~\ref{alg:sap}. 
At the start (line 1), we initialize the parameters of the base-learner $\mathbf{\theta}$ randomly. 
The candidate operation parameters $\mathbf{\phi}$ are initialized to leave the input unaffected (for example, scalars are initialized to 1 and biases to 0). 
The layer-wise activation strengths $w_i^\ell$ of the candidate operations are initialized to the uniform distribution.
After this initialization, we repeat the following steps until a stopping criterion is met, such as having sampled a certain number of task batches, or observing decreasing performance on held-out validation tasks. 
We randomly sample batches of tasks (line 3), initialize the task-specific parameters $\mathbf{\phi}^{(0)}=\mathbf{\phi}$, and make $T$ gradient update steps on the support set of every task (lines 6--8), and perform meta-updates to the initialization parameters $\mathbf{\Theta}$ (line 11) using the query sets of the tasks.
Note that the meta-update requires the computation of second-order gradients as we have to compute the gradient of the inner-level gradients.
The complexity of this is quadratic in the number of parameters, but can be avoided by using the first-order assumption $\nabla_{\mathbf{\phi}} \mathbf{\phi}^{(T)}_j = I$.

\begin{algorithm}
\caption{Subspace Adaptation Prior (SAP)}\label{alg:sap}
\begin{algorithmic}[1]
\Require $p(\Tau)$
\Require $\alpha, \beta$
\State initialize $\mathbf{\theta}, \mathbf{\phi}, \mathbf{\lambda}$
\While{not converged}
    \State sample batch of tasks $\mathcal{B} = \{ \Tau_j = (D^{tr}_{\Tau_j}, D^{te}_{\Tau_j}) \backsim p(\Tau) \}_{j=1}^{M}$
    \For{task $\Tau_j = (D^{tr}_{\Tau_j}, D^{te}_{\Tau_j}) \in \mathcal{B}$}
        \State initialize task-specific parameters $\mathbf{\phi}^{(0)}_j = \phi$
        \For{$t=0,\ldots,T-1$}
            \State compute gradient update $\mathbf{\phi}^{(t+1)}_j$ using Equation~\ref{eq:graddescent} on $D^{tr}_{\Tau_j}$
        \EndFor 
    \EndFor
    \State update initial parameters $\mathbf{\Theta} = \{ \mathbf{\theta}, \mathbf{\phi}, \mathbf{\lambda} \}$ using Equation~\ref{eq:outerobjective}
\EndWhile
\end{algorithmic}
\end{algorithm}

\paragraph{Pruning}
The scores $w^\ell_i$ represent the {\it activation strengths} of the different candidate operations/subspaces, and can also be used for pruning the operations, for example, in a layer-wise or regular top-K fashion. 
By default, we do not hard-prune operations and maintain a convex combination of different candidate operations unless explicitly mentioned otherwise. 
Note that we cannot simply drop low activation strength operations from the network as that changes the composite features and layerwise activation statistics.
Hard-pruning requires re-training the network with only the selected (non-pruned) subspaces/operations.

\subsection{Analysis}
One may wonder what the role is of inserting operation sets $\mathcal{O}^\ell$ in the base-learner network since they have the same expressivity as weight matrices.
In other words, why do we have two consecutive matrix multiplications $\weight^\ell \mathcal{O}^\ell \inp$ if that is equivalent to having one matrix multiplication $\mathcal{U}\inp$, where $\mathbf{U} = \weight^\ell \mathcal{O}^\ell$. 
There are two reasons for maintaining two separate matrices, which we describe below.

\paragraph{Regularization} First, having a set of operations $\mathcal{O}^\ell$ allows SAP to learn which sets, corresponding to weight subspaces of a full-rank matrix, are relevant for a given task distribution. 
Choosing lower-dimensional subspaces is a form of regularization, as fewer parameters can be adjusted to fit the noise in the data. 

\paragraph{Gradient modulation} Second, when computing gradient updates for the operation parameters $\mathbf{\phi}^\ell$ of a given layer $\ell$, the frozen base-layers $\weight^\ell$ implicitly modulate the gradients since the error signal traverses backward through $\weight^\ell$ to $\mathcal{O}^\ell$.
This method of gradient modulation was proposed by \citet{lee2018gradient}. 
Below, we borrow the analysis performed in that paper to illustrate the modulation.

Suppose we are presented with a task $\Tau_j$ and that the output for a given layer in the network is given $\mathbf{v} = \weight \mathcal{O} \inp$, where $\inp$ is the input to the layer.
Moreover, assume that the loss of the network on task $\Tau_j$ is given by $\mathcal{L}_{\Tau_j}$
Then, the parameters of the operations $\mathcal{O}$ are updated using a gradient update step, and we obtain the new output
\begin{align}
    \mathbf{v}^{\mathit{new}} &= \weight (\mathcal{O} - \alpha \nabla_{\mathcal{O}}\mathcal{L}_{\Tau_j})\inp \\
    &=  \mathbf{v} - \alpha (\weight \nabla_{\mathcal{O}}\mathcal{L}_{\Tau_j}) \inp.
\end{align}
Note that we slightly abuse notation here since the parameters of the operations are denoted as $\mathbf{\phi}$. 
As we can see, the change in the layer's output $\Delta ( \mathbf{v}^{\mathit{new}},  \mathbf{v})$ is negatively proportional to the $(\weight \nabla_{\mathcal{O}}\mathcal{L}_{\Tau_j})$.
Here, $\weight$ \emph{warps} the gradients with respect to the operation parameters. 
The warping of these gradients is meta-learned across tasks such that within a few gradient updates in warped space, a good performance can be achieved \citep{Flennerhag2020Meta-Learning, lee2018gradient, parkO19metacurvature}. 

As a consequence, {\it SAP can learn both in which parameter subspaces to perform gradient descent by learning appropriate subsets of operations, as well as learn how to warp these subspaces so that few gradient updates yield good performance.}

\section{Experiments}

In this section, we aim to answer the following research questions:
\begin{itemize}
    \item Does learning suitable layer-wise operations/subspaces improve meta-learning performance on sine wave regression? (Section~\ref{sec:toyregression})
    \item Do the learned strengths of subspaces/operations match the task structure in a simple synthetic setting? (Section~\ref{sec:activsonsine},  Section~\ref{sec:structure})
    \item How well does SAP perform at few-shot image classification? (Section~\ref{sec:fslimage})
    \item How well does SAP perform at cross-domain few-shot image classification? (Section~\ref{sec:crossimagefsl})
    \item Is hard subspace pruning beneficial for the performance of SAP? (Section~\ref{sec:pruning})
    \item What is the influence of second-order gradients on the performance of SAP? (Section~\ref{sec:order})
    \item What operations are important for few-shot image classification? (Section~\ref{sec:learnedspacesimage})
    \item How does SAP compare in terms of the running time and and number of trainable parameters compared to the baselines? (Section~\ref{sec:runtime})
\end{itemize}

\subsection{Sine wave regression}\label{sec:toyregression}

First, we study the few-shot learning performance of SAP on few-shot sine wave regression, which is commonly used in the meta-learning community  \citep{finn2017model, li2017metasgd, parkO19metacurvature}.
Here, the goal is to learn sine wave regression tasks $\Tau_j$ corresponding to sine curves $g_j(x) = A_j\cdot \sin(x-p_j)$ from a limited set of $k$ examples. 
The amplitudes $A_j$ and phases $p_j$ of these sine curves are randomly sampled from the intervals $[0.1, 5.0]$ and $[0, \pi]$, respectively.
While the results on sine-wave regression are not our main contribution, the structure of these problems were a motivation for the development of this method, and therefore this is a good test-case on which we expect SAP to perform well.
Of course, SAP can only be considered a valuable contribution when it also works on more relevant problem types, which we explore in the following sections.  

We use the same base-learner architecture, a fully-connected neural network with 2 hidden ReLU layers of 40 nodes each, as in \citep{finn2017model}.
For the SVD operations (see candidate operations in Section~\ref{sec:sapalgorithm}), we use ranks 5, 10, and 15 in the candidate pools.
All candidate operations were initialized to have no effect on the network predictions at the start (transformation matrices were initialized to identity matrices, biases to 0, and scale operations to 1). 
All techniques are meta-trained on $70\,000$ tasks using one update step per task and a meta-batch size of $4$.
We perform validation every $2\,500$ tasks to select the best performing model, which will be tested after 1 and 10 gradient update steps on $2\,000$ meta-test tasks consisting of $k$ support examples and $50$ query data points.

\begin{table}[htb!]
    \centering
    \begin{tabular}{lrrrrr}
    \toprule
         & & \multicolumn{2}{c}{5-shot} & \multicolumn{2}{c}{10-shot} \\
         \cmidrule(lr){3-4} \cmidrule(lr){5-6}
         & params & T=1 & T=10 & T=1 & T=10 \\
         \midrule
         MAML & $1\,761$ & 0.73 $\pm$ 0.016 & 0.42 $\pm$ 0.011  & 0.49 $\pm$ 0.011 & 0.15 $\pm$ 0.005  \\
         T-Net & $4\,962$ & 0.53 $\pm$ 0.014 & 0.24 $\pm$ 0.009 & 0.33 $\pm$ 0.009 & 0.09 $\pm$ 0.004 \\
         MT-Net & $5\,043$ & 0.55 $\pm$ 0.013 & 0.19 $\pm$ 0.005 & 0.34 $\pm$ 0.008 & 0.06 $\pm$ 0.002 \\
         \midrule
         SAP (ours) & $10\,013$ & \textbf{0.47} $\pm$ 0.012 & \textbf{0.10} $\pm$ 0.003  & \textbf{0.28} $\pm$ 0.008 & \textbf{0.04} $\pm$ 0.001 \\
         \bottomrule
    \end{tabular}
    \caption{The mean MSE meta-test loss on 5- and 10-shot sine wave regression after $T=1$ and $T=10$ update steps. The results are averaged over 5 runs with different random seeds and the 95\% confidence intervals are displayed as $\pm$ x. The number of parameters is shown in the column ``params'', even though the used backbones are equally expressive.}
    \label{tab:sineregression}
\end{table}

As baselines, we compare against MAML, T-Net, and MT-Net \citep{lee2018gradient} as well as Warp-MAML \citep{Flennerhag2020Meta-Learning} as are highly similar to SAP, which allows us to investigate the advantage of SAP's ability to learn which subsets of operations to adjust.
We refrain from comparing against MetaNAS \citep{elsken2020meta}, as this technique also adjusts the architecture at meta-test time and is orthogonal to SAP and the methods we compare against.
For all methods, we use the same hyperparameters as reported in \citep{finn2017model, lee2018gradient}.
In this case, however, Warp-MAML is equivalent to T-Net as both use insert linear ``transformation'' or ``warp'' layers in the base-learner network.  
The results of the experiments are displayed in Table~\ref{tab:sineregression}. 
In this table, we can see that SAP consistently outperforms all tested baselines, supporting the hypothesis that it is indeed beneficial to learn in which subspaces to perform gradient descent. 
We have also performed experiments with SAP and the feature masking method used in MT-Net, where some features are frozen based on learned feature masking probabilities, but found that it decreases the performance, which may be due to the low-dimensional operations present in the architecture, which are more susceptible to being completely frozen as soon as a single feature is masked.

\subsection{The learned subspaces for sine regression}
\label{sec:activsonsine}

\begin{figure}[thb]
    \centering
    \includegraphics[scale=0.4]{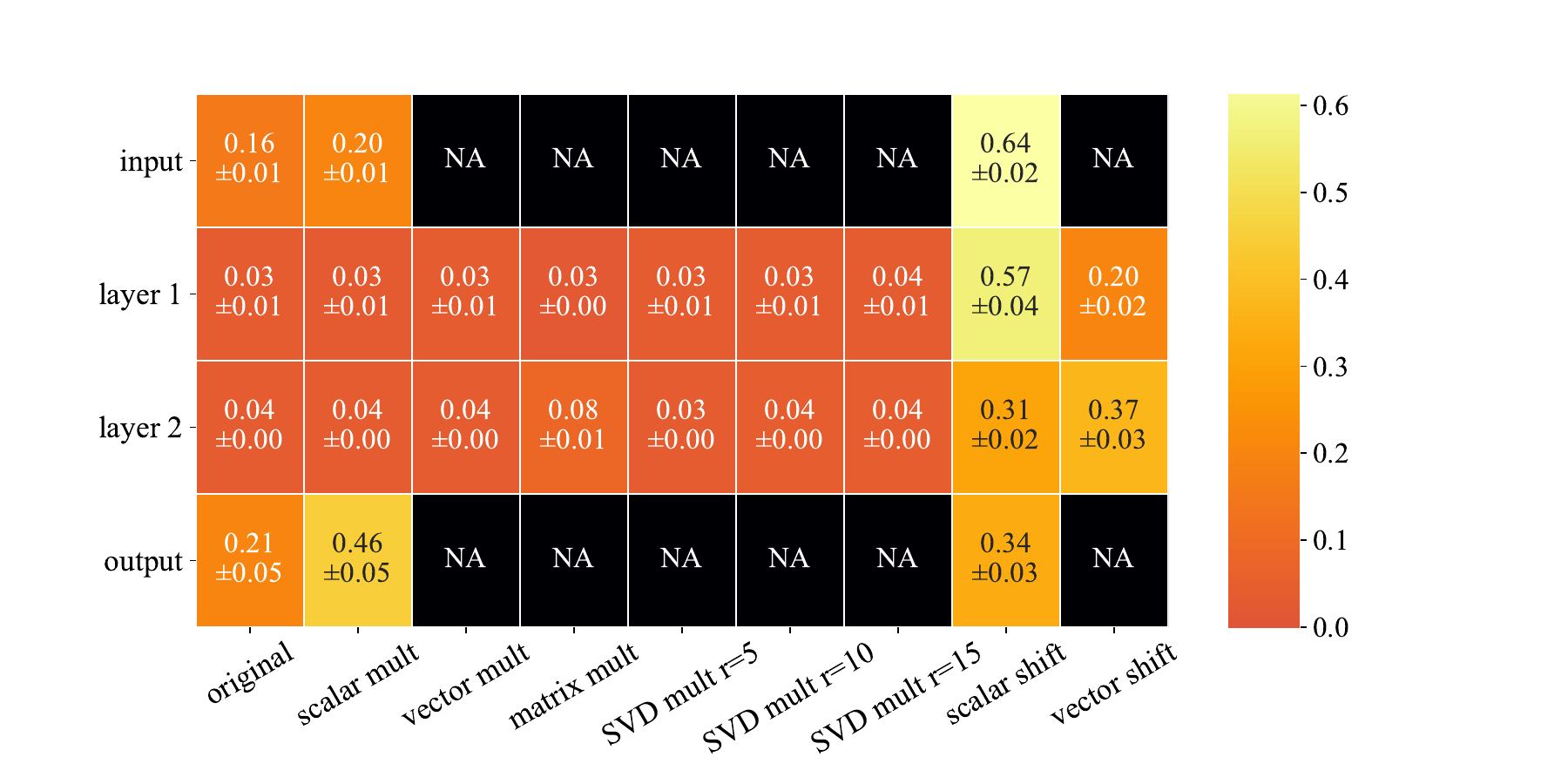}
    \caption{The importance of the different operations in SAP for $5$-shot sine wave regression. The results are averaged across 5 runs with different random seeds and the standard deviations are shown as $\pm x$. NA entries indicate that these operations were not in the candidate pool for that layer, and ``mul'' means multiplication. The y-axis indicates the layer on which the operations act, and the x-axis displays the different candidate operations. Simple scalar multiplication and shifting, and vector shifts obtain high activation strengths in all layers. The input shift and output scale (inherently present in the definition of a sine wave) obtain high activation strengths.}
    \label{fig:sine-alfas-sap}
\end{figure}

Next, we investigate (in the same setting as above) the importance of the different candidate operations for quick adaptation to new tasks to see whether the operations match the task structure.
We hypothesize that shifting the input and scaling the output are important operations as they are inherent in the definition of a sine wave $g_j(x) = A_j\cdot \sin(x-p_j)$. 
To investigate this, we inspect the activation strengths $w^\ell_i$ of the operations of the best models across 5 different runs with different random seeds. 
The operations that were used are were introduced in Table~\ref{tab:pool} (left side). 
The results for SAP with $T=1$ are displayed in Figure~\ref{fig:sine-alfas-sap} (similar results are obtained when making $T=10$ updates and therefore omitted for brevity).
As we can see, the most important transformations on the input and output are a scalar shift and multiplication, respectively. 
In other words, SAP has learned that shifting the input and scaling the output are effective operations to learn new tasks. 
Note that these operations match the structure of sine waves.
While this confirms our hypothesis, SAP also assigns relatively large importance to operations that are not directly observable in the mathematical definition of sine curves such as an output shift and intermediate shifts.

\subsection{Matching the problem structure}
\label{sec:structure}

To further investigate the ability of SAP to match the learned candidate operation strengths to the structure of the problem, we investigate whether changes in the problem structure amount to changes in the learned activation strengths by SAP for the different operations.
For this, we consider a synthetic sine wave regression problem that generalizes the settings studied by \citet{finn2017model} and \citet{li2017metasgd}. 
In this setting, we create different \emph{task families} (task distributions) that are characterized by the mathematical operations inherent in the ground-truth function.
All task families share the following template for the ground-truth function $g(x) = A \cdot \sin(f \cdot x - p) + \beta$, where $A$ is the amplitude, $f$ the frequency, $p$ the phase, and $\beta$ the output offset.
What distinguishes task families is which of these parameters they include in the functional description. 
For example, task family A may fix the amplitude and vary the frequency, phase, and output offset, whereas task family A may vary the amplitude and fix the rest.  
Each task family is thus defined by which of these parameters are varied among tasks from that family and which are kept constant.
If a given parameter is not varied, we fix it to a value that leaves the function unaltered (i.e., $A,f,p=1$ and $\beta=0$).  
In total, there are $2^4 = 16$ task families that can be constructed by varying or fixing these parameters.

\begin{figure}[thb]
    \centering
    \includegraphics[scale=0.3]{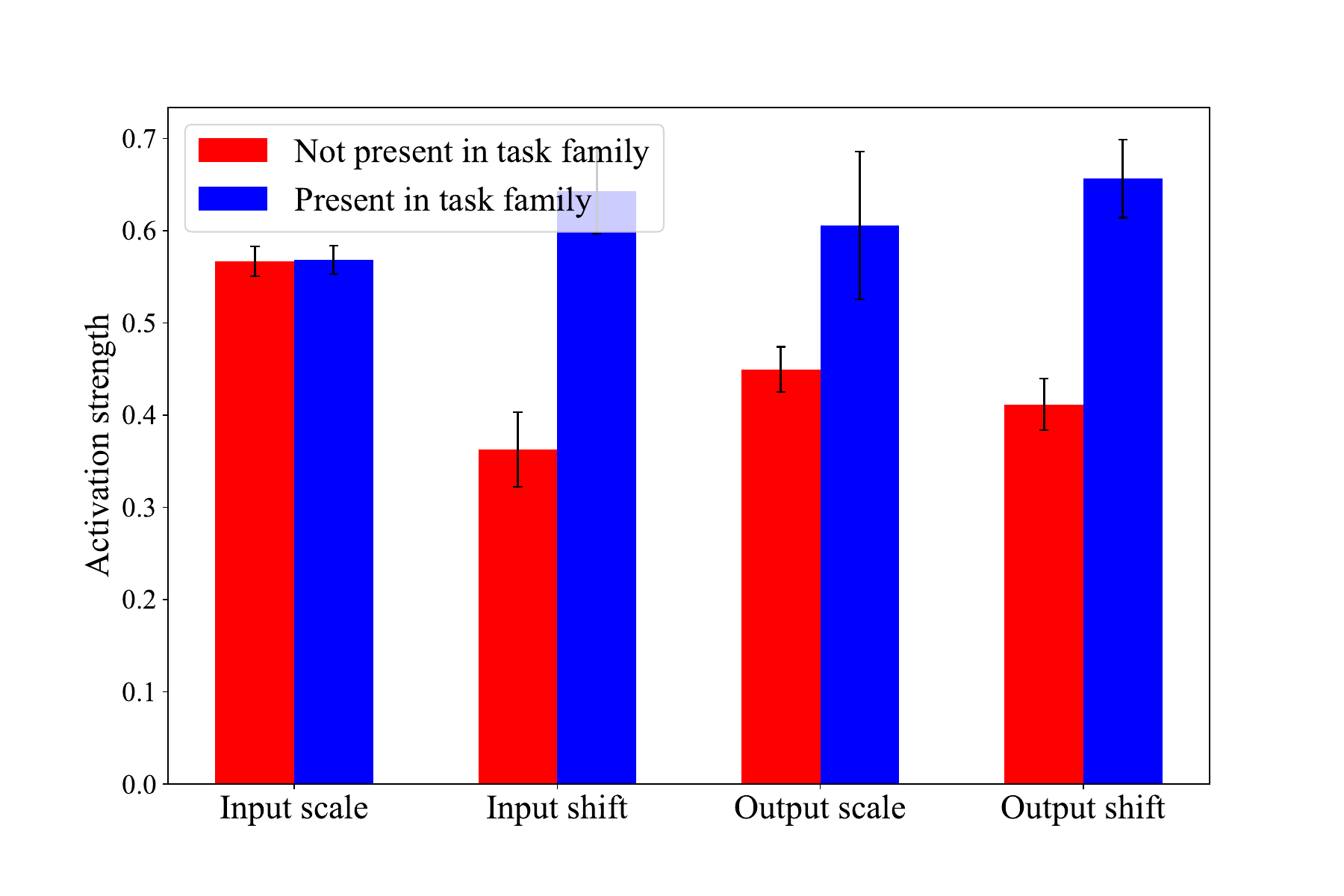}
    \caption{The mean activation strengths of the different operations corresponding to the intrinsic parameters that were varied within task families. The vertical bars display 95\% confidence intervals over 5 runs with different random seeds. Each task family has the following template $g(x) = A \cdot \sin(f \cdot x - p) + \beta$ and differs in which of these operations are varied among tasks. If operations are inherently present in a task family, SAP assigns higher activation strengths to them than if they are not present in 3 out of 4 cases, indicating that the operations often match the problem structure.}
    \label{fig:problstruct}
\end{figure}

We perform meta-training on each of these task families separately and investigate whether SAP discovers the operations that are inherently present in the task structure.
The experimental details follow those used in Section~\ref{sec:toyregression} with the exception that only operations were included that could be present in the task families to be able to measure whether SAP correctly detects and uses them.
We use 20-shots per task and set the number of inner updates to $T=1$.
The results of this experiment are displayed in Figure~\ref{fig:problstruct}. 
As we can see, SAP assigns higher activation strengths to operations that are inherently present in the task families in three out of four cases, i.e., input scale (frequency), input shift (phase), and output shift.
A statistical T-test shows that these differences in mean activation strengths are statistically significant, using a threshold of 0.05. 
For the input scale, however, we observe that SAP assigns similar activation strength to the input scale activation, regardless of whether such an operation was present in the task family. 
This may indicate that SAP uses other operations to compensate for this, such as vector multiplications or matrix multiplications in later layers. 
Overall, these results suggest in this simple synthetic setting, SAP is capable of learning to use operations that appear in the problem structure in 75\% of the scenarios.

\begin{table}[tb!]
    \centering
    \begin{tabular}{ll}
    \toprule 
    Hyperparameter & Range \\
    \midrule
         Inner learning rate & LogUniform(1e-3, 6e-1) \\
         Inner update steps (training) & Uniform(1,10)\\
         Inner update steps (testing) & Uniform(inner steps training, 15)\\
         Meta-batch size & Uniform(1,10) \\
         Gradient masking & Uniform($\{ \text{False}, \text{True} \}$) \\
         \bottomrule 
    \end{tabular}
    \caption{The ranges and sampling types for the hyperparameters, which were tuned with random search. The bounds are inclusive.}
    \label{tab:hypersearchimage}
\end{table}

\subsection{Few-shot image classification}
\label{sec:fslimage}

Next, we investigate the performance of SAP in few-shot image classification settings, where the goal is to learn new image classification tasks from a few examples. 
For this, we use the popular $N$-way $k$-shot classification setup (see Section~\ref{sec:backgroundfsl}) on miniImageNet \citep{vinyals2016matching, ravi2017optimization} and tieredImageNet \citep{ren2018meta}.
We use the frequently used Conv-4 backbone \citep{finn2017model, lee2018gradient, Flennerhag2020Meta-Learning}, consisting of four blocks, where each block contains $3 \times 3$ convolutions, a max pooling layer, 2D BatchNorm, and a ReLU nonlinearity.
In the literature, this backbone has been used with 64 channels for every convolutional block \citep{snell2017prototypical, vinyals2016matching} as well as 32 channels \citep{finn2017model, nichol2018reptile}.
For this reason, we present the results for SAP on both variants. 
The final feature representations are flattened and fed into a softmax output layer. 
All techniques were trained for $60\,000$ episodes and were validated after every $2\,500$ tasks and we use the best-reported hyperparameters by the original authors.

We tuned a subset of the hyperparameters for SAP on the \emph{meta-validation} tasks using random search with a function evaluation budget of 30 runs. 
Each run was restricted to finish within 7 days on a single PNY GeForce RTX 2080TI GPU.
Runs that took longer (e.g., because of a large meta-batch size) were discarded from the hyperparameter search. 
The used hyperparameter ranges and sampling types that were used for the random search are displayed in Table~\ref{tab:hypersearchimage}.
Due to computational constraints, we adopted the best reported hyperparameters of the baseline methods as reported in their respective papers. 
As such, the comparison against these baselines is in this experiment only for illustrative purposes, as the hyperparameter optimization procedure on these methods has not been executed under the same conditions.

\begin{table}[th!]
\centering
\begin{tabular}{lcccccc}
\toprule
& \multicolumn{2}{c}{1-shot} & \multicolumn{2}{c}{5-shot} \\
\cmidrule(lr){2-3} \cmidrule(lr){4-5}
 & 32 channels & 64 channels  & 32 channels & 64 channels \\ 
 \midrule
MAML & 48.0 $\pm$ 0.8 & 46.7 $\pm$ 0.8 & 64.4 $\pm$ 0.4 & 63.6 $\pm$ 0.4 \\
T-Net & 48.9 $\pm$ 0.8 & 48.7 $\pm$ 0.8 & 65.3 $\pm$ 0.4 & - \\
MT-Net & 48.5 $\pm$ 0.8 & 49.3 $\pm$ 0.8 & 63.0 $\pm$ 0.4 & - \\
Warp-MAML & 49.5 $\pm$ 0.8 & 49.8 $\pm$ 0.8 & 63.9 $\pm$ 0.4 & 64.6 $\pm$ 0.4 \\
\midrule
SAP (ours) & \textbf{51.6} $\pm$ 0.8 & \textbf{52.8} $\pm$ 0.8 & \textbf{65.9} $\pm$ 0.4 & \textbf{67.4} $\pm$ 0.4 \\
\bottomrule
\end{tabular}
\caption{Meta-test accuracy scores on 5-way miniImageNet classification over 5 runs with two variants of the Conv-4 backbone, that is, with 32 or 64 channels per block. The 95\% confidence intervals are displayed as $\pm$ x. ``-'' indicates that the experiments required more GPU VRAM than available.}
\label{tab:channelsmin}
\end{table}

The results for the experiments on 5-way miniImageNet and tieredImageNet classification are displayed in Table~\ref{tab:channelsmin} and Table~\ref{tab:channelstiered}.
Note that the results for 5-shot T-Net and MT-Net are missing as they were unable to run on our GPU with 12GB of VRAM. 
As we can see, the performance of the techniques improves when using 64 channels compared with 32, with the exception of MAML on miniImageNet and T-Net in the 1-shot setting on miniImageNet.
As we can see, SAP consistently outperforms all tested baselines in all tested settings (with gains between 1.1\% to 3.3\% accuracy), indicating that it is beneficial to learn subsets of operations on which gradient descent is performed in the case of few-shot image classification. 

\begin{table}[th!]
\centering
\begin{tabular}{lcccccc}
\toprule
& \multicolumn{2}{c}{1-shot} & \multicolumn{2}{c}{5-shot} \\
\cmidrule(lr){2-3} \cmidrule(lr){4-5}
 & 32 channels & 64 channels  & 32 channels & 64 channels \\ 
 \midrule
MAML & 50.7 $\pm$ 0.8 & 51.5 $\pm$ 0.8 & 65.2 $\pm$ 0.4 & 66.6 $\pm$ 0.4 \\
T-Net & 49.4 $\pm$ 0.8 & 51.7 $\pm$ 0.8 & 64.6 $\pm$ 0.4 & - \\ 
MT-Net & 49.8 $\pm$ 0.9 & 51.5 $\pm$ 0.8 & 64.6 $\pm$ 0.4 & - \\
Warp-MAML & 51.8 $\pm$ 0.8 & 53.3 $\pm$ 0.8 & 66.0 $\pm$ 0.4 & 68.2 $\pm$ 0.4 \\
\midrule
SAP (ours) & \textbf{52.9} $\pm$ 0.8 & \textbf{54.5} $\pm$ 0.8 & \textbf{69.3} $\pm$ 0.3 & \textbf{71.3} $\pm$ 0.4 \\
\bottomrule
\end{tabular}
\caption{Meta-test accuracy scores on 5-way tieredImageNet classification over 5 runs with two variants of the Conv-4 backbone, that is, with 32 or 64 channels per block. The 95\% confidence intervals are displayed as $\pm$ x. ``-'' indicates that the experiments required more GPU VRAM than available. }
\label{tab:channelstiered}
\end{table}

\subsection{Cross-domain few-shot image classification}
\label{sec:crossimagefsl}

Next, we study the performance of SAP in a more challenging \emph{cross-domain} few-shot image classification setting.
In this setting, techniques are trained on tasks from dataset $A$ and evaluated on tasks from another dataset $B$, in contrast to the setting used above, where the techniques were evaluated on \emph{unseen} tasks from the same dataset used for training. 
We use the same setting as \citet{chen2019closer}, in which we train on miniImageNet and evaluate on CUB \citep{wah2011caltech}.
In addition, we also train on tieredImageNet \citep{ren2018meta} and test on CUB. 
All other experimental details are the same as above. 

The results of this experiment are shown in Table~\ref{tab:crossdomain}.
As we can see, SAP performs on par with Warp-MAML in the 1-shot setting for MIN $\rightarrow$ CUB.
Both outperform the other tested baselines in that scenario. 
In other cases, however, SAP yields performance improvements ranging from 0.5\% to 3.9\% accuracy. 
This supports the hypothesis that it is beneficial to learn which subsets of operations to adjust when learning new tasks. 

\begin{table}[th!]
\centering
\begin{tabular}{lcccccc}
\toprule
& \multicolumn{2}{c}{MIN $\rightarrow$ CUB} & \multicolumn{2}{c}{Tiered $\rightarrow$ CUB} \\
\cline{2-3} \cline{4-5}
 & 1-shot & 5-shot  & 1-shot & 5-shot \\ 
\midrule
MAML &  37.3 $\pm$ 0.3 & 54.7 $\pm$ 0.3 & 38.1 $\pm$ 0.3 & 55.1 $\pm$ 0.3 \\
T-Net & 38.0 $\pm$ 0.3 & 55.6 $\pm$ 0.3 & 37.5 $\pm$ 0.3 & 54.8 $\pm$ 0.3 \\
MT-Net & 37.1 $\pm$ 0.3 & 53.1 $\pm$ 0.3 & 38.0 $\pm$ 0.3 & 55.5 $\pm$ 0.3 \\
Warp-MAML & \textbf{41.0} $\pm$ 0.3 & 55.3 $\pm$ 0.3 & 40.9 $\pm$ 0.3 & 56.8 $\pm$ 0.3 \\
\midrule
SAP (ours) &  40.9 $\pm$ 0.3 & \textbf{55.8} $\pm$ 0.3 & \textbf{41.1} $\pm$ 0.3 & \textbf{60.7} $\pm$ 0.3 \\
\bottomrule
\end{tabular}
\caption{Average cross-domain meta-test accuracy scores over 5 runs a 32-channel Conv-4 backbone. Techniques trained on tasks from one data set were evaluated on tasks from another data set. The 95\% confidence intervals are displayed as $\pm$ x.}
\label{tab:crossdomain}
\end{table}

\subsection{Effect of hard pruning}
\label{sec:pruning}
 
Next, we investigate the effect of hard pruning the number of operations per layer, which is a common feature of DARTS~\citep{liu2018darts}, and therefore also inherited by SAP.
For this, we compare the performance of SAP without hard pruning and SAP where we only retain the top-K operations as indicated by their strength scores.
The hard-pruned SAP is re-trained using only the candidate operations which were not pruned. 
The results of this experiment with a 32-channel Conv-4 backbone are displayed in Table~\ref{tab:pruning} (for the 64-channel variant, please see Table~\ref{tab:crossdomain64channels} in the appendix).
As we can see, hard pruning can have a mild positive effect on the meta-learning performance, whilst reducing computational costs due to the fact that fewer parameters have to be trained.
This also implies that some operations may indeed be suboptimal for a given task distribution, which soft-pruning is not able to completely filter out, and that a model which fully excludes these, can achieve better performance. 
We note, however, that the 95\% confidence intervals are overlapping, suggesting that these performance increases are not significant.

\begin{table}[tbh!]
\centering
\begin{tabular}{lcccc}
\toprule
& \multicolumn{2}{c}{miniImageNet} & \multicolumn{2}{c}{tieredImageNet} \\
\cmidrule(lr){2-3} \cmidrule(lr){4-5}
 & 1-shot & 5-shot & 1-shot & 5-shot \\ 
\midrule
No pruning & 51.6 $\pm$ 0.8 & 65.9 $\pm$ 0.4 & 52.9 $\pm$ 0.8 & 69.3 $\pm$ 0.3 \\
\midrule
Top-1 & 51.4 $\pm$ 0.8 & 65.8 $\pm$ 0.4 & 52.8 $\pm$ 0.8 & 69.4 $\pm$ 0.4  \\
Top-2 & \textbf{51.8} $\pm$ 0.8 & \textbf{66.3} $\pm$ 0.4 & \textbf{53.4} $\pm$ 0.8 & 69.4 $\pm$ 0.4  \\
Top-3 & \textbf{51.8} $\pm$ 0.8 & \textbf{66.3} $\pm$ 0.4 & 53.0 $\pm$ 0.9 & 69.9 $\pm$ 0.4  \\
\bottomrule
\end{tabular}
\caption{Mean meta-test accuracy scores on miniImageNet and tieredImageNet with 95\% confidence intervals over 5 different runs. We used a Conv-4 backbone with 32 channels for these results.}
\label{tab:pruning}
\end{table}

\subsection{The effect of the gradient order}
\label{sec:order}

All tested techniques require the computation of second-order gradients by default. 
Here, we investigate how the performance of SAP is affected by making a first-order approximation.
We compare this first-order variant with the regular second-order variant, using the same experimental settings as used in Section~\ref{sec:fslimage}.
The results of this experiment are shown in Table~\ref{tab:order}. 
As we can see, the first-order approximation is consistently outperformed by the regular variant, with differences between 0.2\% and 7.3 \% accuracy, indicating that second-order gradients play an important role in achieving good performance. 

\begin{table}[th!]
\centering
\begin{tabular}{lcccccc}
\toprule
& \multicolumn{2}{c}{miniImageNet} & \multicolumn{2}{c}{tieredImageNet} \\
\cmidrule(lr){2-3} \cmidrule(lr){4-5}
 & 1-shot & 5-shot  & 1-shot & 5-shot \\ 
 \midrule
SAP (first-order) &  51.4 $\pm$ 0.8 & 63.7 $\pm$ 0.4 & 47.2 $\pm$ 0.8 & 62.0 $\pm$ 0.4 \\
SAP (second-order) & \textbf{51.6} $\pm$ 0.8 & \textbf{65.9} $\pm$ 0.4 & \textbf{52.9} $\pm$ 0.8 & \textbf{69.3} $\pm$ 0.3 \\
\bottomrule
\end{tabular}
\caption{Meta-test accuracy scores on miniImageNet and tieredImageNet classification over 5 runs using the Conv-4 backbone with 32 channels. The 95\% confidence intervals are displayed as $\pm$ x.}
\label{tab:order}
\end{table}

\subsection{The learned subspaces for image classification}
\label{sec:learnedspacesimage}

\begin{figure}[thb]
    \centering
    \includegraphics[scale=0.4]{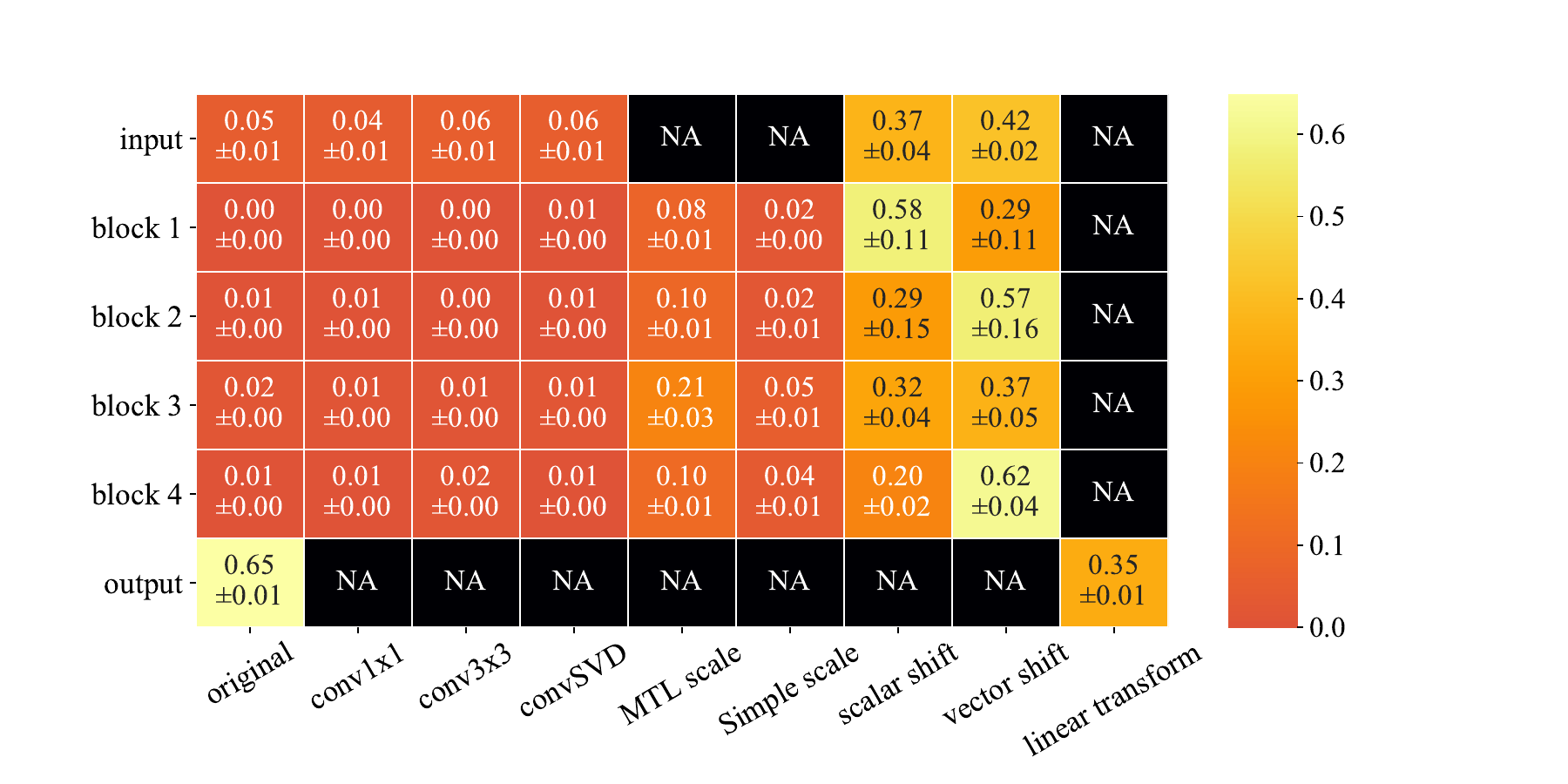}
    \caption{ The importance of the different subspaces/operations in SAP on 5-way 1-shot miniImageNet using Conv-4 with 32 channels. The results are averaged across 5 runs with different random seeds and the standard deviations are shown as $\pm x$. NA entries indicate that these operations were not in the candidate pool for that layer. Simple scalar shift and vector shift operations obtain the highest activation strengths throughout the convolutional network.}
    \label{fig:conv4}
\end{figure}

In order to gain insight into what operations are important for achieving good few-shot learning performance in SAP, we investigate the learned activation strengths for the different candidate operations.
The operations that were used are were introduced in Table~\ref{tab:pool} (right side). 
In Figure~\ref{fig:conv4}, we can see these learned strengths in SAP on 1-shot 5-way miniImageNet using the Conv-4 backbone with 32 channels (similar patterns are seen for the backbone with 64 channels as can be seen in Figure~\ref{fig:conv464channels} in the appendix). 
As we can see, high-dimensional convolutional operations (conv1x1, conv3x3, convSVD) obtain low activation strengths, while lower-dimensional subspaces/operations such as shifts (scalar and vector) and MTL scale yield larger strengths.
The greatest strength is assigned to the former throughout all layers.
This may indicate that the higher-dimensional operations lead to overfitting, while the lower-dimensional operations are more suited for adapting to tasks when only limited data is available. 
Consequently, this implies that it is indeed beneficial to adapt subsets of operations when learning new tasks.

\subsection{Number of parameters and running time}\label{sec:runtime}

Lastly, we compare the running times and the number of parameters used by the different methods on few-shot image classification.
These statistics were measured whilst performing the experiments in Section~\ref{sec:fslimage} and the results are displayed in Table~\ref{tab:time}. 
As we can see, SAP has the largest number of parameters, even though the backbone is equally expressive as that used by others.
The running time of SAP, however, is often less than that of the baselines. 
This is caused by the fact that all methods use different hyperparameter settings in order to optimize the performance, which relates to the running time. 
For example, a larger meta-batch size or number of updates per task leads to an increase in running time. 
SAP uses the smallest meta-batch size and number of updates and hence yields the quickest running time.
Note that these runtimes do not include the hyperparameter optimization that was performed, which adds a factor to the runtimes.

\begin{table}[th!]
\centering
\begin{adjustbox}{width=\linewidth}
\begin{tabular}{lrrrrr}
\toprule
&  & \multicolumn{2}{c}{miniImageNet} & \multicolumn{2}{c}{tieredImageNet} \\
\cmidrule(lr){3-4} \cmidrule(lr){5-6}
 & params & 1-shot & 5-shot  & 1-shot & 5-shot \\ 
 \midrule
MAML & $32\,901$ & 11h36min $\pm$ 7min & 8h20min $\pm$ 1min & 11h34min $\pm$ 5min & 8h25min $\pm$ 4min \\
T-Net & $37\,022$ &33h05min $\pm$ 19min & 30h20min $\pm$ 7min & 33h25min $\pm$ 23min & 30h30min $\pm$ 17min \\ 
MT-Net & $37\,150$ & 33h33min $\pm$ 8min & 30h22min $\pm$ 14min & 33h49min $\pm$ 40min & 30h47min $\pm$ 18min \\
Warp-MAML & $60\,645$ &7h19min $\pm$ 6min  & 7h17min $\pm$ 8min & 7h35min $\pm$ 13min & 7h19min $\pm$ 5min \\
SAP (first-order) & $106\,196$ &1h26min $\pm$ 2min & 1h4min $\pm$ 0min & 1h51min $\pm$ 6min &  2h12min $\pm$ 4min \\ 
SAP &  $106\,196$ & 3h59min $\pm$ 0min & 6h09min $\pm$ 0min & 1h51min $\pm$ 6min & 9h24min $\pm$ 19min  \\
\bottomrule
\end{tabular}
\end{adjustbox}
\caption{The number of trainable parameters (``params'') and mean running times on miniImageNet and tieredImageNet classification over 5 runs using the Conv-4 backbone with 32 channels. The standard deviations are displayed as $\pm$ x min. In spite of the differences in the number of parameters, the backbones are equally expressive. SAP was found to work best with a small meta-batch size and number of updates per task compared with the other approaches and hence yields the quickest running time.}
\label{tab:time}
\end{table}

\section{Conclusions}

In this work, we introduced, \emph{Subspace Adaptation Prior} (SAP), a novel meta-learning algorithm that jointly learns a good neural network initialization and good parameter subspaces (or subsets of operations) in which new tasks can be learned within a few gradient descent updates from a few data. 
SAP overcomes the limitations of current state-of-the-art gradient-based meta-learning techniques which perform gradient descent in \emph{full parameter space} as they adjust all parameters \citep{finn2017model, lee2018gradient, Flennerhag2020Meta-Learning}, which may be suboptimal, and may lead to overfitting during few-shot learning.
Note, however, that our goal is not to yield state-of-the-art performance. 
Instead, we investigate the question of whether the few-shot learning performance of deep neural networks can be improved by meta-learning which subsets of parameters to adjust.

Our experiments show that SAP outperforms similar existing gradient-based meta-learners in few-shot sine wave regression, yields better performance in single-domain few-shot image classification settings, and yields competitive or superior performance in cross-domain few-shot image classification. 
This highlights the advantage of learning suitable subspaces in which to perform gradient descent when learning new tasks.
This could be due to the regularization effect of not having to adjust all parameters as well as due to the ability to match structures inherently present in task families.
Our experiments in Section~\ref{sec:structure} on synthetic task families demonstrate that the SAP is able to learn operations that match the task structure in simple settings in 75\% of the cases.
In other cases, it may compensate by using other operations that are not inherently present in the task structure.

Inspection of the subspace activation strengths in few-shot image classification reveals that simple and low-dimensional operations, such as shifting features by a single scalar or element-wise by a vector, are important.
This is in line with recent work and findings \citep{triantafillou2021learning, requeima2019fast, bateni2020improved} which show that adapting pre-trained embeddings by means of such low-dimensional transformations, such as FiLM layers \citep{perez2018film}, can yield excellent performance.  
Furthermore, we found that hard-pruning the subspaces in SAP, or operations, such that only a discrete subset is used instead of a convex combination, was slightly beneficial, although no statistically significant differences were found.   

\paragraph{Future work}

One limitation of SAP is that it requires the computation of second-order gradients by default during meta-training in order to update the initialization parameters, in a similar fashion as other gradient-based meta-learners such as MAML \citep{finn2017model}, (M)T-Net \citep{lee2018gradient}, and Warp-MAML \citep{Flennerhag2020Meta-Learning}. 
These second-order gradients require $O(N^2)$ storage, where $N$ is the number of total network parameters, which is prohibitive for deep networks. 
This limitation can be bypassed by using a first-order approximation, which comes at the cost of a performance penalty (between 0.2\% and 7.3\% accuracy in our experiments).

Gradient-based meta-learning methods struggle to scale well to deep networks as recent work suggests that simple pre-training and fine-tuning of the output layer \citep{tian2020rethinking, chen2021meta, huisman2021preliminary} can yield superior performance on common few-shot image classification benchmarks. 
This is also the reason, besides searching for energy-efficient few-shot learners, that in our experiments we focus on relatively shallow backbones that adapt all layers when learning new tasks, instead of only the output layer.

Other limitations are that SAP introduces more parameters and that the candidate pools of operations are selected by hand, despite the fact that these operations are general.
One direction for future work could be to design a method to discover such subspaces from scratch, instead of relying on a candidate set of operations, perhaps using an auto-encoder that generates the weights of a layer based on latent codes as used by \citet{rusu2018meta}.
Masking the adaptation of these latent codes using Gumbel-softmax \citep{jang2017gumbel, maddison17} as done by MT-Net \citep{lee2018gradient} would amount to adjusting only a subset of the parameters when performing gradient descent.
This can reduce the number of parameters and may also help to scale gradient-based meta-learners, including SAP, to deep networks and make them competitive with approaches relying on pre-trained features, which is an open challenge.

Finally, orthogonal work has proposed a method that can also adjust the architecture during the meta-test phase~\citep{elsken2020meta}. 
Since this showed great potential, it would be worthwhile to combine this with SAP. 
Moreover, it would be interesting to investigate the sensitivity of SAP related methods such as MetaNAS to the chosen operations or blocks that these methods can select to use. 
We leave these ideas for future work, which has the potential to further advance the state-of-the-art.

\section*{Acknowledgments}

This work was performed using the compute resources from the Academic Leiden Interdisciplinary Cluster Environment (ALICE) provided by Leiden University.

\section*{Declarations}

\subsection*{Funding}
Not applicable: no funding was received for this work.

\subsection*{Conflict of interest} All authors certify that they have no affiliations with or involvement in any organization or entity with any financial interest or non-financial interest in the subject matter or materials discussed in this manuscript.

\subsection*{Ethics approval}
Not applicable. 

\subsection*{Consent to participate}
Not applicable.

\subsection*{Consent for publication}

Not applicable: this research does not involve personal data, and publishing of this manuscript will not result in the disruption of any individual's privacy.

\subsection*{Availability of data and material}

All data that was used in this research have been published as benchmarks by \citet{deng2009imagenet,vinyals2016matching} (miniImageNet), \citet{ren2018meta} (tieredImageNet) and \citet{wah2011caltech} (CUB), and is publicly available. The data generator for sine wave regression experiments can be found in the provided code (see below).

\subsection*{Code availability}
All code that was used for this research is made publicly available at \url{https://github.com/mikehuisman/subspace-adaptation-prior}.

\subsection*{Authors' contributions}

MH has conducted the research presented in this manuscript.
AP and JvR have regularly provided feedback on the work, contributed towards the interpretation of results, and have critically revised the whole.    

All authors approve the current version to be published and agree to be accountable for all aspects of the work in ensuring that questions related to the accuracy or integrity of any part of the work are appropriately investigated and resolved.

\subsection*{Employment} All authors declare that there is no recent, present, or anticipated employment by any organization that may gain or lose financially through the publication of this manuscript.

{
\bibliographystyle{abbrvnat}
\bibliography{sn-bibliography}
}

\appendix

\label{app:exps}

\section{Additional experimental results}

In this appendix, we show additional experimental results on few-shot image classification. 

\subsection{Validation of re-implementation}

\begin{table}[thb!]
\centering
\begin{tabular}{lcccccc}
\toprule
& \multicolumn{2}{c}{1-shot} & \multicolumn{2}{c}{5-shot} \\
\cmidrule(lr){2-3} \cmidrule(lr){4-5}
 & Reported & Local Repr  & Reported & Local repr \\ 
\midrule
MAML        & 48.7 $\pm$ 1.8 & 48.0 $\pm$ 0.8 & 63.2 $\pm$ 0.9 & 64.4 $\pm$ 0.4 \\
T-Net & 50.9 $\pm$ 1.8 & 48.9 $\pm$ 0.8 & - & 65.3 $\pm$ 0.4  \\
MT-Net & 51.7 $\pm$ 1.8  & 48.5 $\pm$ 0.8 & - & 63.0 $\pm$ 0.4  \\
Warp-MAML$^*$ & -  & 49.5 $\pm$ 0.8 & - & 63.9 $\pm$ 0.4 \\
\midrule
SAP (ours) & - & \textbf{51.6} $\pm$ 0.8 & - & \textbf{65.9} $\pm$ 0.4  \\
\bottomrule
\end{tabular}
\caption{Mean meta-test accuracy scores on 5-way miniImageNet classification over 5 runs using a Conv-4 backbone with 32 channels. The 95\% confidence intervals are displayed as $\pm$ x. $^*$ \citet{Flennerhag2020Meta-Learning} only reported the performance of Warp-MAML with 128 feature maps per convolutional block instead of 32, as displayed in the table.}
\label{tab:imagefsl}
\end{table}

We re-implemented the baselines to ensure a fair comparison in the used setting, and because the code of Warp-MAML has not been made available for other researchers. 
To verify our re-implementations of the baselines (T-Net, MT-Net, and Warp-MAML), we compare the reported performances to the ones that we obtain.
The results of the image classification experiments are displayed in Table~\ref{tab:imagefsl}.
As we can see, there are minor differences between the reported performances and our local reproduction of their results.
Also with the original code of T-Net and MT-Net, we were unable to reproduce their results.
Other people have encountered similar issues reproducing the reported numbers of meta-learning techniques, including MAML, T-Net, and MT-Net.\footnote{There is an open issue on the GitHub repository of MT-Net about the inability to reproduce their reported results on miniImageNet. See \url{https://github.com/yoonholee/MT-net/issues/5}. Other researchers such as \citet{antoniou2018how} have also reported issues reproducing MAML.}

\subsection{Cross-domain few-shot image classification}
In Table~\ref{tab:crossdomain64channels}, we show the cross-domain few-shot learning classification results when using 64 channels with the Conv-4 backbone. 
Also in this case, SAP outperforms other tested baselines. 
We also note that the performance of SAP is improved when using 64 channels compared with 32 (see Section~\ref{sec:crossimagefsl}). 

\begin{table}[th!]
\centering
\begin{tabular}{lcccccc}
\toprule
& \multicolumn{2}{c}{MIN $\rightarrow$ CUB} & \multicolumn{2}{c}{Tiered $\rightarrow$ CUB} \\
\cmidrule(lr){2-3} \cmidrule(lr){4-5}
 & 1-shot & 5-shot  & 1-shot & 5-shot \\ 
\midrule
MAML &  37.1 $\pm$ 0.3 & 53.7 $\pm$ 0.3 & 38.8 $\pm$ 0.3 & 56.8 $\pm$ 0.3 \\
T-Net & 38.3 $\pm$ 0.3 & OOM & 39.9 $\pm$ 0.3 & OOM \\
MT-Net & 37.3 $\pm$ 0.3 & OOM & 39.1 $\pm$ 0.3 & OOM \\
Warp-MAML & 40.7 $\pm$ 0.3 & 56.2 $\pm$ 0.3 & 42.5 $\pm$ 0.3 & 58.9 $\pm$ 0.3 \\
\midrule
SAP (ours) &  41.6 $\pm$ 0.3 & 57.8 $\pm$ 0.3 & 43.3 $\pm$ 0.3 & 64.3 $\pm$ 0.3 \\
\bottomrule
\end{tabular}
\caption{Average cross-domain meta-test accuracy scores over 5 runs using a 64-channel Conv-4 backbone. Techniques trained on tasks from one data set were evaluated on tasks from another data set. The 95\% confidence intervals are displayed as $\pm$ x.}
\label{tab:crossdomain64channels}
\end{table}

\subsection{The effect of hard pruning}

Table~\ref{tab:pruning64} displays the effect of hard pruning when using 64 channels instead of 32. 
As we can see, hard pruning is slightly beneficial, but again, not significantly. 

\begin{table}[tbh!]
\centering
\begin{tabular}{lcccc}
\toprule
& \multicolumn{2}{c}{miniImageNet} & \multicolumn{2}{c}{tieredImageNet} \\
\cmidrule(lr){2-3} \cmidrule(lr){4-5}
 & 1-shot & 5-shot & 1-shot & 5-shot \\ 
\midrule
No pruning & 52.8 $\pm$ 0.8  & 67.4 $\pm$ 0.4 & 54.5 $\pm$ 0.8 & 71.3 $\pm$ 0.4 \\
\midrule
Top-1 & 52.8 $\pm$ 0.8 & \textbf{67.6} $\pm$ 0.4 & \textbf{55.1} $\pm$ 0.8 & \textbf{72.7} $\pm$ 0.4  \\
Top-2 & \textbf{52.9} $\pm$ 0.8 & \textbf{67.6} $\pm$ 0.4 & 54.1 $\pm$ 0.8 & \textbf{72.7} $\pm$ 0.4 \\
Top-3 & 52.6 $\pm$ 0.8 & 67.4 $\pm$ 0.4 & 55.0 $\pm$ 0.8 & 72.4 $\pm$ 0.4  \\
\bottomrule
\end{tabular}
\caption{Mean meta-test accuracy scores on 5-way miniImageNet and tieredImageNet classification with 95\% confidence intervals computed over 5 different runs. We used a Conv-4 backbone with 64 channels for these results.}
\label{tab:pruning64}
\end{table}

\subsection{The learned subspaces for image classification}

\begin{figure}[tb]
    \centering
    \includegraphics[scale=0.4]{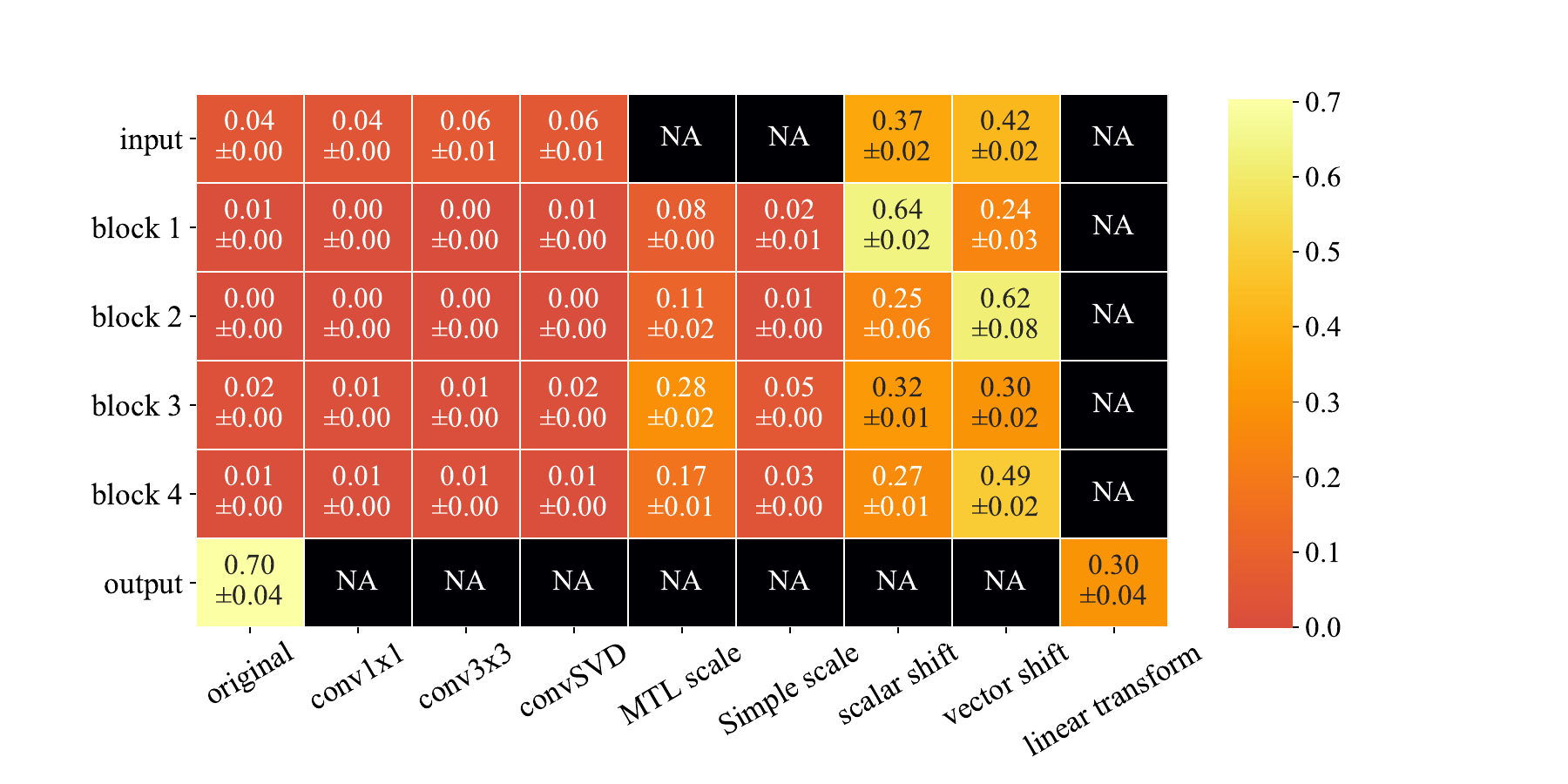}
    \caption{ The importance of the different subspaces/operations in SAP on 5-way 1-shot miniImageNet using Conv-4 with 64 channels. The results are averaged across 5 runs with different random seeds and the standard deviations are shown as $\pm x$. NA entries indicate that these operations were not in the candidate pool for that layer. Simple scalar shift and vector shift operations obtain the highest activation strengths throughout the convolutional network.}
    \label{fig:conv464channels}
\end{figure}

Figure~\ref{fig:conv464channels} displays the learned activation strengths of SAP on 5-way 1-shot miniImageNet using Conv-4 with 64 channels. 
Similar patterns are observed for the 32-channel case.

\end{document}